\documentclass[journal]{IEEEtran}
\usepackage{graphicx}
\usepackage{float}
\usepackage{siunitx}
\usepackage[table,xcdraw]{xcolor}
\usepackage{nameref}
\usepackage[caption=false]{subfig}
\usepackage{amsmath}
\usepackage{amssymb}
\usepackage[ruled,vlined]{algorithm2e}
\usepackage{multirow}
\usepackage[super]{nth}
\usepackage{booktabs}
\usepackage{authblk}

\usepackage{xcolor}
\usepackage[normalem]{ulem}

\usepackage[pagebackref=false,breaklinks=true,letterpaper=true,colorlinks,bookmarks=false]{hyperref}
 
\DeclareMathOperator{\EXMEAN}{\mathbb{E}}
\DeclareMathOperator{\BATCH}{\mathcal{B}}

\newcommand{\midsepremove}{\aboverulesep = 0mm \belowrulesep = 0mm}
\newcommand{\midsepdefault}{\aboverulesep = 0.605mm \belowrulesep = 0.984mm}
\newcommand{\etal}{\textit{et al}. }
\newcommand{\thickhline}{%
    \noalign {\ifnum 0=`}\fi \hrule height 1pt
    \futurelet \reserved@a \@xhline
    }
\usepackage{graphicx}
\graphicspath{{pics/}{figs/}}

\begin{document}

\title{Stabilizing Inputs to Approximated Nonlinear Functions for Inference with Homomorphic Encryption in Deep Neural Networks}

\author[1]{Moustafa AboulAtta}
\author[ ]{Matthias Ossadnik}
\author[2]{Seyed-Ahmad Ahmadi}
\affil[1]{Technische Universit{\"a}t M{\"u}nchen}
\affil[2]{German Center for Vertigo and Balance Disorders\\Ludwig-Maximilians Universit{\"a}t M{\"u}nchen}

%



\maketitle

\begin{abstract}
		Leveled Homomorphic Encryption (LHE) offers a potential solution that could allow sectors with sensitive data to utilize the cloud and securely deploy their models for remote inference with Deep Neural Networks (DNN). However, this application faces several obstacles due to the limitations of LHE. One of the main problems is the incompatibility of commonly used nonlinear functions in DNN with the operations supported by LHE, i.e. addition and multiplication. As common in LHE approaches, we train a model with a nonlinear function, and replace it with a low-degree polynomial approximation at inference time on private data. While this typically leads to approximation errors and loss in prediction accuracy, we propose a method that reduces this loss to small values or eliminates it entirely, depending on simple hyper-parameters. This is achieved by the introduction of a novel and elegantly simple Min-Max normalization scheme, which scales inputs to nonlinear functions into ranges with low approximation error. While being intuitive in its concept and trivial to implement, we empirically show that it offers a stable and effective approximation solution to nonlinear functions in DNN. In return, this can enable deeper networks with LHE, and facilitate the development of security- and privacy-aware analytics applications.
		
	\end{abstract}
	\section{Introduction}
	Deep Neural Networks (DNN) have achieved remarkable success across different domains. One of the factors that led to the spread of these technologies in industry is the emergence of cloud service providers. It lowered the entry cost to the usage of DNN by offering different payment schemes. However, some sectors like the medical, financial or governmental can have constraints that prevent them from using these services due to the sensitivity of their data. The risk of exposing a patient's data to a third party due to loss or theft has serious ethical implications and could have serious legal or ethical ramifications.\\
	A potential solution that can allow these sectors to utilize cloud services is offered by \textit{Leveled (Fully) Homomorphic Encryption (LHE)}. LHE is a class of encryption schemes that enables computations directly on the ciphertexts without decrypting them first. It can be used to apply DNN models on encrypted data. In this scenario the data owner, e.g. a clinic, would deploy its trained model on the cloud and send the encrypted patient's data as input to the model. Then it will receive the predictions back which are still encrypted. The data owner can then minimize the risk of losing or exposing the data to any other party including the service provider.\\
	Unfortunately, LHE has several limitations that prevents direct porting of DNN models used in the literature and the industry to usage with private data. The two main obstacles are:
\begin{enumerate}
\item Mathematical operations are restricted to addition and multiplication. Consequently, it is impossible to use the nonlinear functions that are commonly used like ReLU \cite{Nair2010}.
\item With each mathematical operation, the noise level in the ciphertext  grows and can silently corrupt the message if it grows beyond a certain threshold. This limits the possible depth of a neural network.
\end{enumerate}
    In literature (see \autoref{sec:related_work}), two general approaches have been proposed to solve this limitation: 
\begin{enumerate}
\item Train and infer with polynomial approximations of the nonlinear functions.
\item Train with the original function then replace it with the polynomial approximation at inference time.
\end{enumerate}
    The first approach is difficult to train due to problems like exploding gradient. The second suffers from considerable loss in predictions accuracy as the network gets deeper due to the accumulation of approximations errors (see \autoref{sec:related_work}).\\
    In the following, we propose an improvement to the second approach.  We identify the validity range of polynomial approximations as a major driver of inaccuracies, and introduce a novel Min-Max normalization scheme that restricts function inputs to ranges with low approximation error. Empirical results suggest that this scheme results in a significant reduction of prediction accuracy loss, especially as the network gets deeper, and even when low-degree polynomial approximations are employed.
    
	\section{Preliminaries}
	In this section, we revisit a few aspects of LHE that are relevant to this work. We give an overview of the most common components used in DNN, focusing on the contrast between commonly employed DNN functions and approximations supported by LHE schemes.
	
	\subsection{Homomorphic Encryption}\label{sec:homomorphic_enc}
	The first \textit{fully homomorphic encryption} (FHE) scheme was devised by Gentry in his dissertation \cite{Gentry2009}. To allow the computation of an arbitrary number of operations, he proposed a procedure called \textit{bootstrapping} that resets the noise to its initial level after each operation. However, it was computationally too expensive for industrial use cases. Building on his work, several other schemes were developed that skipped the usage of bootstrapping and thus allowed faster computation, but only up to a certain depth, after which the ciphertext gets corrupted. This branch of schemes is usually called LHE. 
	Among LHE schemes, there have been different developments, each with their own advantages and limitations, like BGV \cite{Brakerski2014}, BFV \cite{Fan2012} and HEAAN \cite{Cheon2017}. We briefly introduce those schemes with focus on relevant properties to this work. 
	
	\paragraph{Supported operations}
	LHE schemes support only the basic mathematical functions \textit{addition} and \textit{multiplication}. The operands could be a combination of ciphertexts and plaintexts. LHE implementations (e.g. SEAL \cite{seal23}) offer optimizations for operations based on those two functions, like exponentiation by a non-private constant. From here on out, any other function has to be replaced or approximated using the two aforementioned functions. Additionally, BGV, BFV and HEAAN support vectorization by offering the ability to pack several messages into a single ciphertext and exploit the Single Instruction Multiple Data (SIMD) paradigm \cite{Gentry2012}.
	
	\paragraph{Noise growth and depth of mathematical circuits}\label{para:noise_growth}
	Noise is an inherent part of a ciphertext in LHE. Even a newly encrypted ciphertext will have some noise, which is responsible of the probabilistic property of LHE. With each application of a mathematical function, the noise increases. When the noise level exceeds a certain threshold, the ciphertext is corrupted and the encrypted message becomes irretrievable. This noise threshold is controlled with different hyper-parameters at the initialization time of the crypto-system. However, attempts at increasing this threshold will result in more expensive memory and computation costs, hence, it is beneficial to keep the depth of the mathematical circuit as low as possible. In particular, the noise generated by multiplication of two ciphertexts is a lot greater than multiplication of a ciphertext with a plaintext or additions with any combination, hence, it is common in literature to associate the noise growth directly with the private multiplicative depth of a circuit.
	
	\paragraph{Integers and rational numbers} 
	In most applications and especially in DNN, the values of inputs are real numbers, however, most LHE schemes including BGV and BFV do not support floating point numbers. Most users resort to two approaches of fixed-point representation of values in a given message. 
	The first approach is to scale the numbers by a $10^k$ factor to take into account $k$ digits after the decimal point and neglect the remainder, then encode the outcome as integers. The advantage of this approach is the simplicity of implementation, however, it requires the user to keep track of the factoring along the arithmetic circuit, which needs to be undone after decryption to get the correct result. Additionally, since division is not supported, the encrypted numbers will keep growing, which requires an initialization of a crypto-system with higher memory and computational costs.
	The second approach is to preserve part of the ciphertext to the integral part and another to the fractional part of the original message. This approach has the advantage that division is possible through the multiplication by a fraction. However, vectorization is not possible anymore as it is only supported with integers. We refer the user to section 3 in \cite{Dowlin2017} for further details.
		
	\subsection{Key Functions in Neural Networks}\label{sec:neural_networks}
	The focus on this work is on artificial feed-forward neural networks. They are built by cascading several layers of operations in sequence, hence, they can be thought of as a mathematical circuit, where each layer increases the depth. Usually, these networks are visualized as a stack with the input to the network as the bottom layer. Below, we list commonly used neural network layers, with an emphasis on the operations that they use. In terms of nomenclature, nonlinear layers are sometimes called activation layers in literature. For clarity, in the remainder of this paper, we will refer to the output of a layer as the activations of that layer, and to the nonlinear layers as such.
	
	\paragraph{Fully connected and/or convolutional layers} The function applied in these components is a \textit{weighted sum}, i.e. the input vector is multiplied by a weights vector, added to a bias vector and then the results are summed. Since these layers constitutes only addition and multiplication, they are fully supported by LHE.
	
	\paragraph{Nonlinear activation layers} These layers are usually applied directly after fully connected or convolutional layers, giving the network the needed nonlinear flexibility to model complex functions. Some of the most common are the logistic function, hyperbolic tangent and the ReLU. LHE cannot support the first two due the existence of exponentiation (and division when integer encoding is used), and it cannot support ReLU due to the conditional branching. Hence, these functions need to be approximated or replaced by others that only employ addition and multiplication. 

	\paragraph{Pooling layers} In DNNs, pooling layers are periodically incorporated to increase the receptive field, to reduce the spatial size of the data, to minimize the amount of parameters and consequently limit the possibility of overfitting, as well as to reduce the memory and computational cost. Two of the most commonly used pooling layers are maximum pooling and average pooling. LHE does not support the maximum function, whereas averaging can be employed if fractional encoding is used as explained in \autoref{sec:homomorphic_enc}.

	\section{Related Work}\label{sec:related_work}
	\subsection*{Neural Networks with Homomorphic Encryption}
	The problem of applying deep neural networks with LHE is gaining momentum in the research community. Related literature in this field covers multiple aspects, like prediction accuracy or computational efficiency. To keep the paper concise, we focus only on the work that directly relates to ours i.e. the work on nonlinear functions.\\
	CryptoNets \cite{Dowlin2016} first demonstrated the application of neural networks on homomorphically encrypted data. The network was based on integers, prohibiting the usage of the division operation. Instead, authors proposed to use sum pooling (called scaled average pooling in the paper) instead of max or average pooling. The network had two nonlinear layers, which utilized the square function as the nonlinearity. It was applied on MNIST dataset \cite{LeCunYann1998} and achieved an accuracy of 99\%. However, as the authors noted, the square function is not a good nonlinear function because it causes an explosion of the gradient as the network gets deeper.\\
	Hesamifard \etal built on CryptoNets in \cite{Hesamifard2016} and \cite{Hesamifard2017} and proposed an alternative to the square function. Their solution was to generate approximations of the nonlinear functions. To keep errors low, the authors optimized the approximation function precisely within the range of expected inputs to the nonlinearity, which required a sampling of the distribution of input values. The authors focused on the sigmoid function. Even though they reported good performance, their solution is challenging to implement and train, as the network architect has to progressively build the network to accommodate those distributions, and generate new approximations for each new layer and for each new dataset or architecture.\\
	Chabanne \etal \cite{Chabanne2017} propose adding \textit{batch normalization} \cite{Ioffe2015a} before each nonlinear layer, which removes the mean of the distribution of activations. They then suggest to generate approximation functions with low error around the origin. They experimented with MNIST dataset on a network with 6 nonlinear ReLU functions. Similar to our approach, they train with the original ReLU nonlinearity, and replace it with the polynomial approximation at inference time. However, since batch normalization does not constrain a distribution to lie within a certain range, some inputs to the nonlinearity could lie in areas with very high approximation error and deviate the network from the correct prediction. To mitigate this problem, the authors propose an extra training phase after replacing the nonlinear functions with the approximations, treating the coefficients of the polynomial as learnable parameters. However, reported performances still had deficiencies, especially when the more desirable low-degree approximations were applied.
	
	\subsection*{Robustness of Neural Networks}
	The topic of neural network robustness to perturbations or lost precision in the activation values is an active field of research. It has direct implications on topics like resiliency against adversarial examples \cite{Yuan2017} or training and inference with limited resources \cite{Gupta:2015}. Even though those works do not directly tackle the challenges faced when utilizing homomorphic encryption, their findings are applicable to those challenges.\\
	Cheney \etal \cite{CheneySK17} developed three methods to evaluate the robustness of a neural network. One of which is the addition of random \textit{weight perturbations} which has the same effect on the signal in the network as the perturbations generated from the nonlinear function approximations in our work. Empirical results suggest that the last layers in the network are relatively resilient to perturbations, whereas the early layers were much more fragile.\\
	Boura \etal \cite{BouraGG18} added some random error to the output of each activation in each nonlinear layer. The authors report that the experiments on all networks that were tested proved resilient to at least 10\% error, with little impact on the global accuracy.
	
	\section{Methods}
	In this section, we present i) the methodology of generating the polynomial approximations of the nonlinear functions and ii) we describe our proposed method for constraining the range of inputs to those approximations, with the aim of keeping approximation errors as low as possible.
	
	\subsection{Approximation of nonlinear functions} \label{sec:act_fn}
\begin{figure}
		\centering
		\includegraphics[width=1\linewidth]{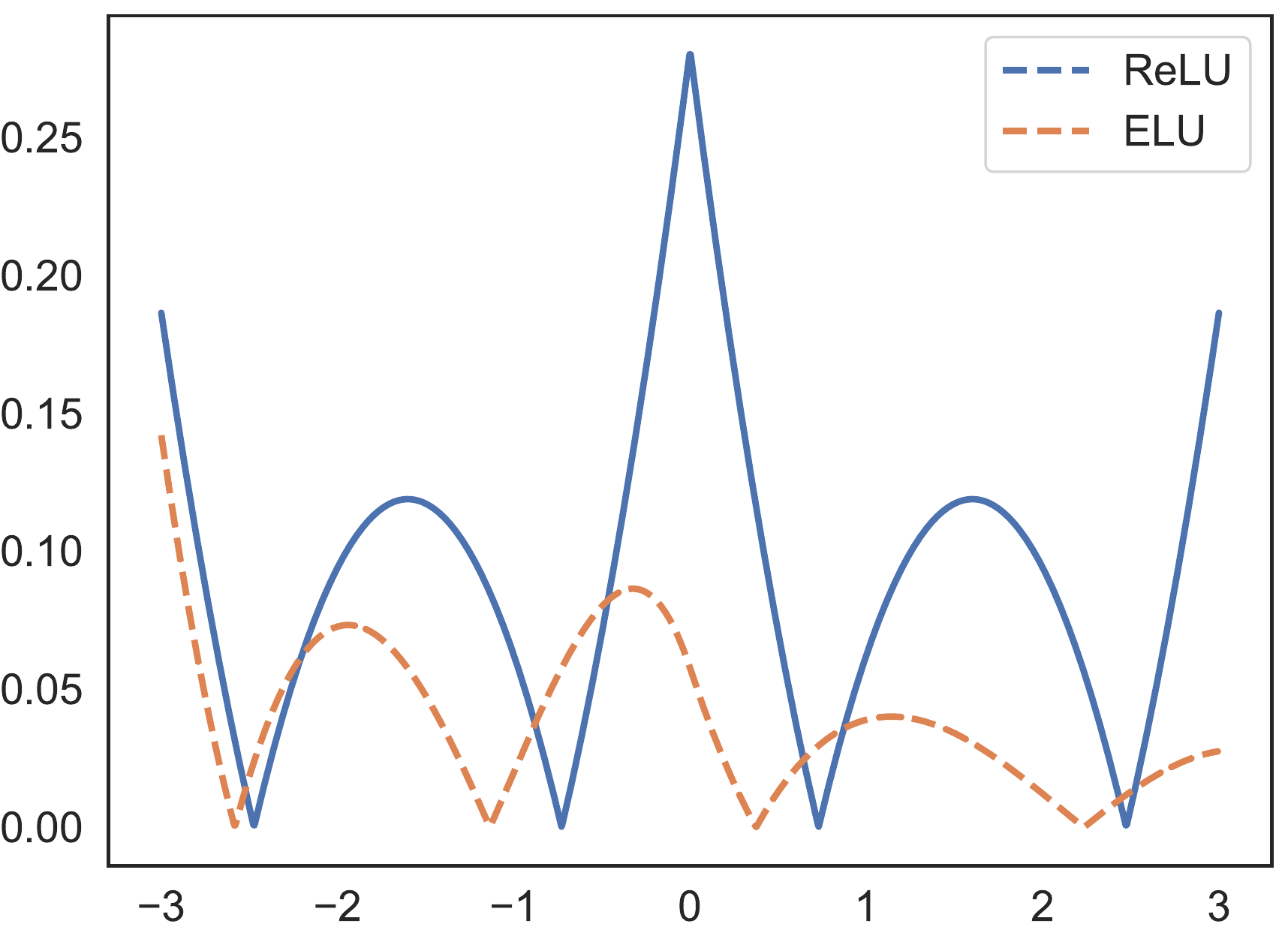}
		\caption{The absolute approximation error of the original nonlinear functions with their corresponding polynomial approximation of degree 3 and approximation range $[-3, 3]$. The high approximation error around the origin in ReLU is noticeable.}
		\label{fig:act_fn_abs_err}
	\end{figure}

\begin{figure*}
        \centering          
        \subfloat[]{\includegraphics[width=0.49\linewidth]{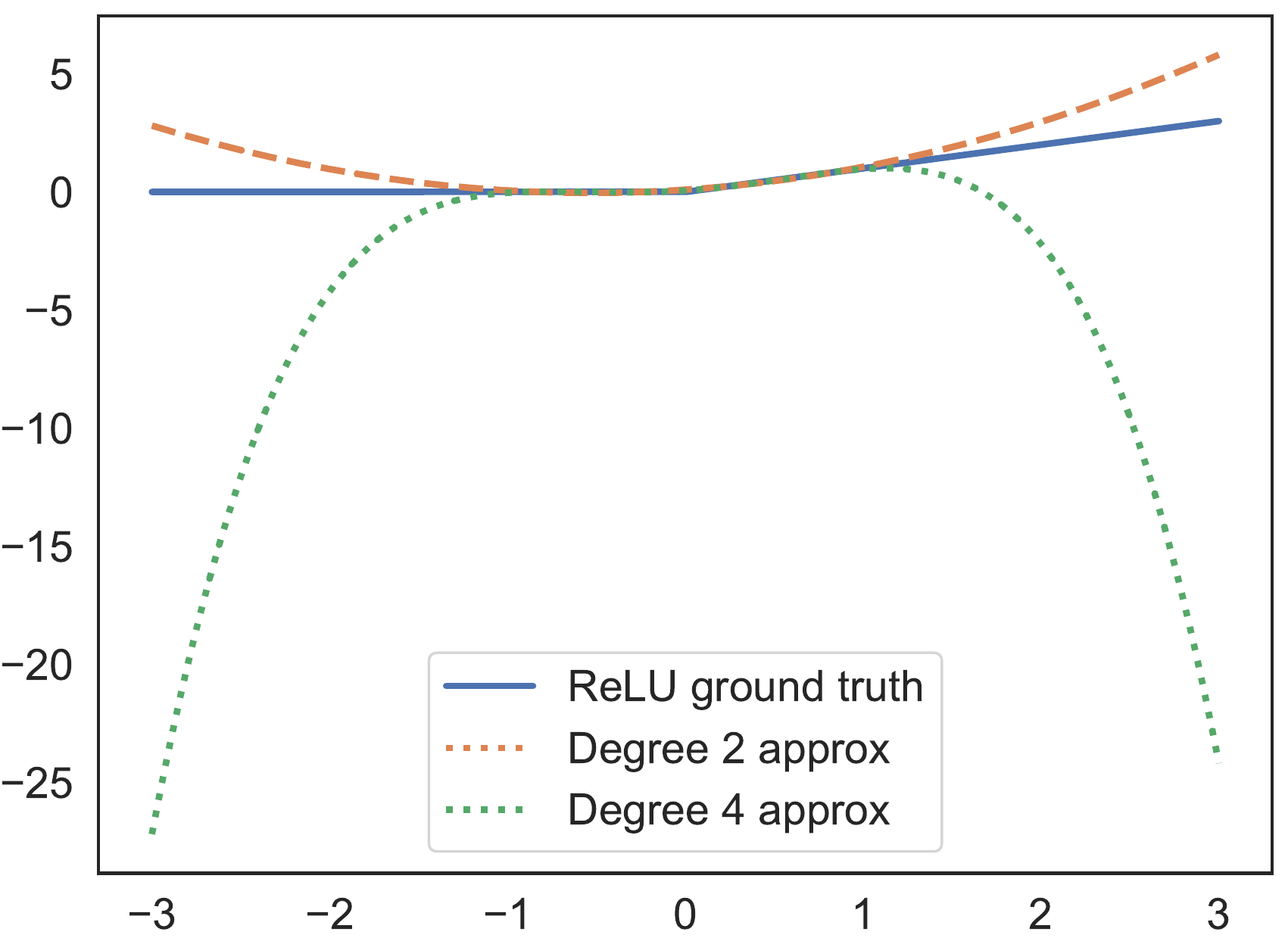}}             
        \subfloat[]{\includegraphics[width=0.49\linewidth]{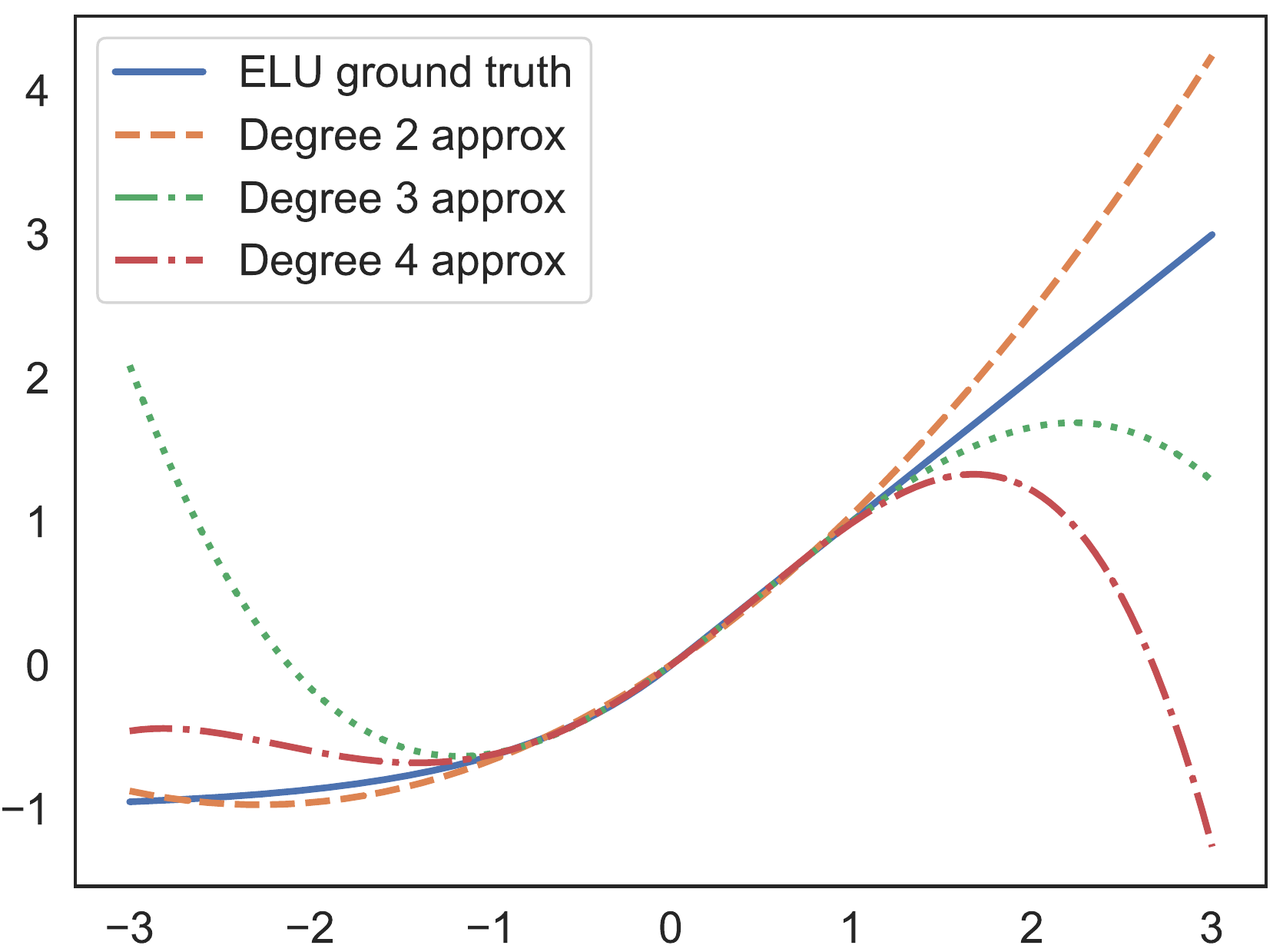}}         
        \caption{Ground truths of the nonlinear functions and their polynomial approximations of different degrees approximated in the range $[-1, 1]$. (a) ReLU and (b) ELU.}%
	\label{fig:fn_approximations}%
	\end{figure*}

	Since LHE supports addition and multiplication, polynomial functions are good candidates for approximation of the nonlinear functions. As stated by the Stone\textendash Weierstrass theorem \cite{hunter2001} \cite{stone1948}, given a sufficient polynomial degree, any continuous function defined on a closed interval can be approximated as a polynomial. However, when polynomials are used in the context of DNN with LHE, the generation of the approximations has to balance between:
    \begin{enumerate}
        \item Low approximation errors, to minimize the propagation of errors and preserve the prediction accuracy.
        \item Low polynomial degrees, to minimize the noise generation in the ciphertext.
    \end{enumerate}
    In our experiments, we utilize Chebyshev polynomials as an orthogonal basis for approximation, while keeping the degree as low as possible between 2 and 6.\\
	We experiment with ReLU due to its widespread usage in modern DNN. However, low degree approximations of ReLU suffer from relatively high approximation errors around the discontinuity at the 0 origin. Consequently, we also experiment with the Exponential Linear Unit (ELU) \cite{Clevert2015}, which is smooth and can be approximated with lower errors. A comparison between the absolute error of the two functions is visualized in \autoref{fig:act_fn_abs_err}. To generate the approximation, we fit the Chebyshev series of the desired degree in a least-squares manner to points drawn from a uniform distribution centered at 0, in order to preserve the nonlinearity of the original functions. Notably, outside the approximation range, the polynomial tends to deviate greatly from the ground truth nonlinearities. See \autoref{fig:fn_approximations}.

	\subsection{Min-Max Normalization Layer}\label{sec:minmax}
	We propose a \textit{Min-Max normalization layer} that aims at strictly keeping the inputs to the polynomial functions within the approximation range. It transforms the inputs to the \textit{target scaling range}, by normalizing the inputs into values within the range $[0,1]$, and then scaling it to the approximation range. More formally:
	 
	\begin{equation}\label{eq:minmax}
	x_{norm}= \underbrace{(r_{max} - r_{min})}_\text{scaling} * \underbrace{\frac{x - x_{min}}{x_{max} - x_{min}}}_\text{[0,1] norm} + \underbrace{r_{min}}_\text{shifting}
	\end{equation}
	 
	where $x$ is the input, $x_{min}$ and $x_{max}$ are the minimum and maximum values in the input respectively. $r_{min}$ and $r_{max}$ are the target range boundaries, which are definable by the network architect and can be intuitively set to the same range used to approximate the original nonlinear functions.\\
	Ideally, the minima and maxima would be computed on the whole input at each layer, however, this is not possible when mini-batches are used in stochastic gradient training. Since parameter updates are applied after each training step, this would turn the previous extrema irrelevant. Consequently, each mini-batch is normalized using minima and maxima in the current mini-batch as can be seen in \autoref{alg:minmax}. For convolutional neural networks, the normalization is applied with respect to feature maps, in order to preserve the convolutional property.\\
	At inference, the moving mean of minima and maxima computed during the training steps is used. This leads to a deterministic model where the prediction of a single sample depends only on its own features and is independent from other unseen samples that are inferred at the same time. This means that at inference time, this layer becomes simply a linear transformation with addition, multiplication and division, which makes it compatible with the LHE implementation that support division through fractional encoding (ex. SEAL\cite{seal23}).\\
	\begin{algorithm}
		\KwIn{Mini-batch of size $m$: $\BATCH=\{x_1...x_m\}$; target range: $r_{min}$ and $r_{max}$}
	 	\KwOut{Normalized  and scaled $\BATCH$}
		$x_{\BATCH, min} \gets \min(\BATCH)$\\
	 	$x_{\BATCH, max} \gets \max(\BATCH)$\\
	 	$\BATCH_{norm} \gets \frac{\BATCH - x_{\BATCH, min}}{x_{\BATCH, max} - x_{\BATCH, min}}$\\
	 	$\BATCH_{scaled} \gets (r_{max} - r_{min})*\BATCH_{norm} + r_{min}$
	 	\caption{{\bf Min-Max normalization during training} \label{alg:minmax}}	
	\end{algorithm}
	To make this layer compatible with other LHE implementations that do not support division, the linear transformation can be easily absorbed into the weights of the preceding convolutional/dense layer. For the convolutional layers, after training the model, the weights are updated as follows:
	\begin{equation*}
	 	w\prime = \frac{(r_{max}-r_{min})w}{\EXMEAN(x_{max}) - \EXMEAN(x_{min})}
 	\end{equation*}
 	\begin{equation*}
 	 	b\prime = \frac{(r_{max}-r_{min})(b-\EXMEAN(x_{min}))}{\EXMEAN(x_{max}) - \EXMEAN(x_{min})}
 	\end{equation*}
	where $w$ and $b$ are the original weights and bias. $\EXMEAN(x_{min})$, $\EXMEAN(x_{max})$ are the moving mean of minima and maxima.
	
	Similar to batch normalization \cite{Ioffe2015a}, this layer has a regularization effect since each training example's features are seen in conjunction and in relative values to the other examples in the same mini-batch. Hence, the model does not produce deterministic values for a single training example (only for identical batches), which was found to improve generalization by Ioffe and Szegedy \cite{Ioffe2015a}.
	
	\section{Experiments}\label{sec:experiments}
	The Min-Max normalization layer is tested with two different baseline networks with different complexities:
	\paragraph{LeNet-5-like and MNIST}	
	A LeNet-like architecture \cite{LeCun1998} is applied on the MNIST dataset \cite{LeCunYann1998}. To reduce the number of parameters, and accordingly the number of the computationally and memory expensive homomorphic evaluations, we follow the Network in Network \cite{Lin2013} approach and replace the fully connected layers with convolutional layers. The output layer has 10 feature maps, corresponding to 10 target classes, followed by a global average pooling layer or global sum pooling layer to obtain a division free network. Altogether, the network has five hidden convolutional layers.
		\paragraph{SqueezeNet and CIFAR-10}
	To the best of our knowledge, a scaled down version of SqueezeNet \cite{Iandola2016, Dathathri2018} is the deepest homomorphically evaluated network in the literature as of the writing of this paper. That version was optimized for the CIFAR-10 dataset \cite{Krizhevsky2009} with a total of 10 nonlinear functions. In this work, we use the original SqueezeNet with a total of 18 nonlinear functions. Importantly, a homomorphical evaluation of this configuration is so far not possible, since other limitations like noise growth are still active fields of research. Parallel research is tackling such limitations, e.g by Chen \etal \cite{chen2018}, who investigate a novel encryption scheme that enables the usage of mathematical circuits up to depth 9, with the same initialization that would only allow a circuit of depth 2 with BFV. In parallel, we consider our work as an important effort to improve nonlinear function approximations in very deep LHE networks, and demonstrate for the first time the feasibility of using up to 18 nonlinear functions without a significant loss in accuracy.
	
	In both networks, LHE-incompatible max pooling was replaced with average pooling as the subsampling layer. We use Keras \cite{Chollet2015a} to train the model with a Tensorflow framework \cite{TensorFlow2017} backend. For all experiments, Adadelta \cite{Zeiler2012} is used as the optimizer and categorical crossentropy as the loss function.\\
	Each network has two baseline models: 
	\paragraph*{Baseline 1} is the network with the original nonlinear function, trained without the Min-Max layer, i.e. a configuration that would be used if no LHE scheme was applied.
	\paragraph*{Baseline 2} is the network with the original nonlinear function, trained with the Min-Max layer. This is the model where the nonlinear approximations will be plugged in place of the original nonlinear functions for LHE inference, and it is used to measure the impact of adding Min-Max layers to the model, compared to Baseline 1.
	
	It is important to note that our main concern during the experiments was the robustness of the network to the perturbations created by the approximated functions as the number of those functions increase. Hence, our attention was directed towards the loss in accuracy in comparison to the baseline networks and not to the absolute accuracy.

	\begin{figure*}
		\centering
		\subfloat[]{{\includegraphics[width=0.45\linewidth]{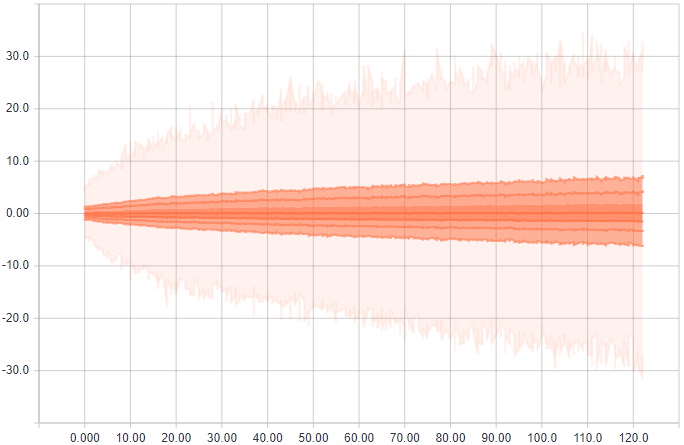} }}%
		\subfloat[]{{\includegraphics[width=0.45\linewidth]{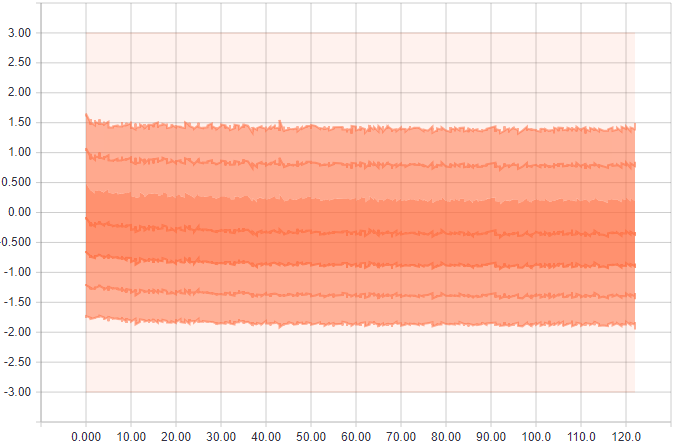} }}%
		\caption{The output values of block 5 in SqueezeNet during training, (a) without deploying the Min-Max layer in the network and (b), with Min-Max normalization layer in range $[-3, 3]$. Notice the stability of the activations which are passed to the nonlinear function within the desired range, when Min-Max is used in contrast to the widening distribution of activations when it is not used. Plots were generated by TensorBoard\cite{TensorBoard2017}.}%
		\label{fig:activations_dists}%
	\end{figure*}
	
	\begin{figure}
		\centering
		\includegraphics[width=1\linewidth]{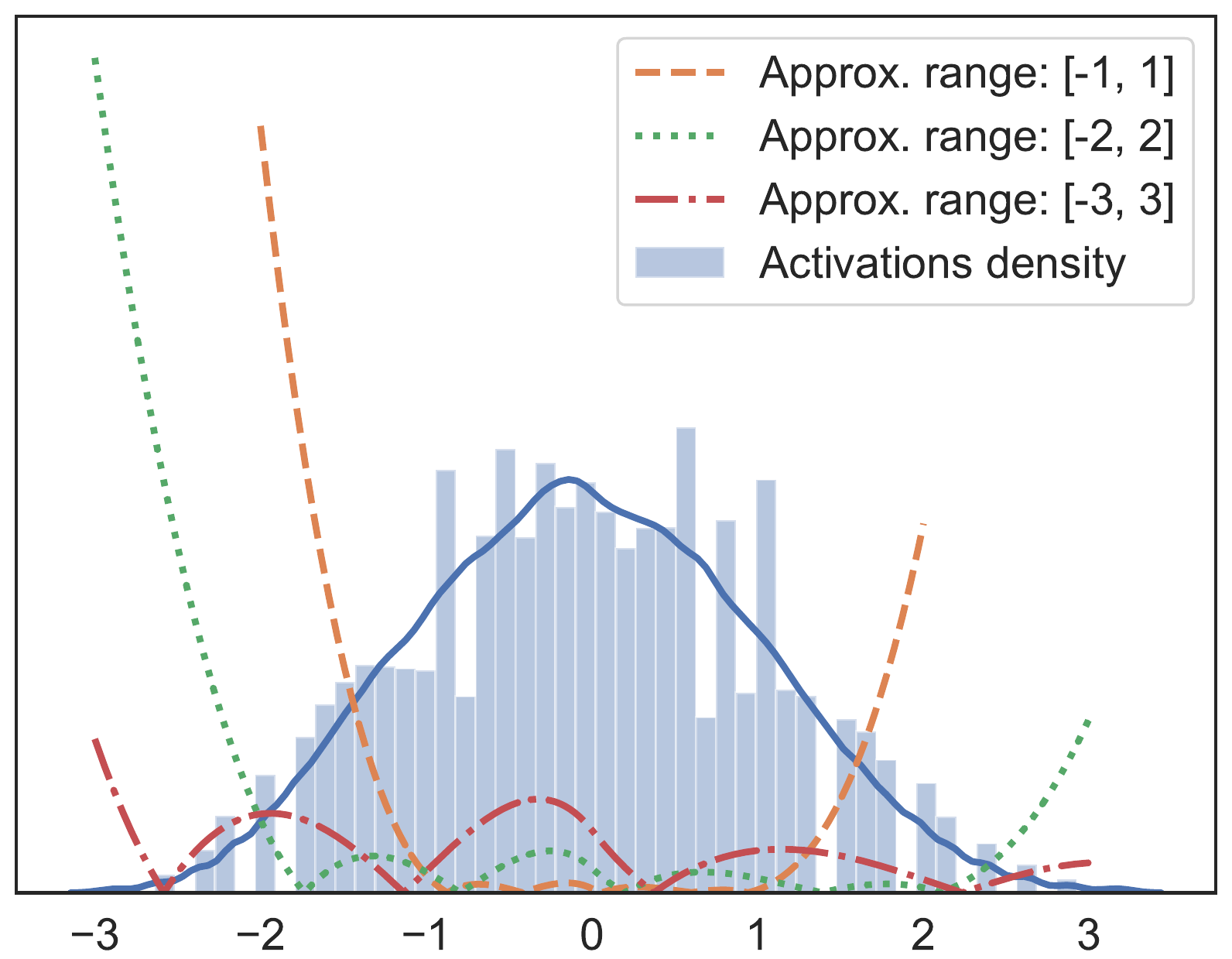}
		\caption{Activations in block 5 in SqueezeNet during a forward pass on an unseen example, right before applying the nonlinear function (in blue) and the approximation error magnitude of approximations of ELU with different approximation ranges (dashed lines in red, green and orange).}
		\label{fig:error_dist}
	\end{figure}

    After evaluating the baselines, each network was initialized with different values of the hyperparameters of the proposed method:
    \begin{itemize}
        \item the \textit{target scaling range} in the Min-Max layer was set to one of the ranges $[-1,1]$, $[-2, 2]$ or $[-3, 3]$.
        \item the \textit{approximation polynomial degree} was varied from 2 to 6 for ELU, and to 2, 4 or 6 for ReLU.
        \item the \textit{approximation range} was varied from $[-1, 1]$ up to the target scaling range.
    \end{itemize}
    
    Using the moving mean of extrema on the training batches showed close accuracy in estimating the extrema of the whole training set and results in normalization of the unseen examples within the desired range as can be seen in \autoref{fig:error_dist}. The blue histogram visualizes the distribution of an unseen example's activation values of the same layer visualized in \autoref{fig:activations_dists}. 

    The behaviour of the hyperparameters was consistent across the two architectures, independent of the network complexity. The target scaling range did not have a significant impact on the performance, as can be seen in the performance of baseline~2 in contrast to the vanilla baseline~1 in \autoref{table:results}. An exception is marked by ELU when the target scaling range is between $[-1, 1]$. The LeNet-like architecture's accuracy decreased by 1.23 percentage points from baseline 1, and by 1.39 and 1.61 points using target scaling ranges of $[-2, 2]$  and $[-3, 3]$ respectively. With SqueezeNet, the accuracy dropped by 0.99 points compared to baseline~1, and by 8.02 and 5.59 from target scaling ranges of $[-2, 2]$  and $[-3, 3]$ respectively (see rows 1-4 in \autoref{table:results}).
    
    For low values of polynomial degrees, low values of approximation range performed best. This behaviour was consistent even when the employed approximation range did not match and was smaller than the target scaling range. For example, when a target scaling layer of range $[-3, 3]$ was employed, a polynomial approximation of \nth{2} degree and approximation range of $[-2, 2]$ performed better than a matching degree and approximation range of $[-3, 3]$. However, as the polynomial degree increased, the polynomial function with the approximation range that matched the target scaling range performed best. Additionally, with the exception when the target scaling range was $[-1, 1]$, ELU's approximations constantly achieved or exceeded the performance of ReLU's approximations with a lower polynomial degree. For example, for SqueezeNet with a target scaling range $[-3, 3]$,  ELU's approximation achieved a loss of 0.34 points with degree 3, whereas ReLU's approximation with degree 4 achieved a loss of 1.75 points (see rows 5 and 6 in \autoref{table:results}).
	
	\begin{table}
		\centering
		\caption{The accuracy of each of the baseline activation functions and their corresponding approximations in the networks with Min-Max scaling layer. $n$ is the polynomial degree and $r$ is the approximation range. The missing values indicate that the best approximation of the given degree is the same as the approximation of the lower degree, e.g. the best approximations for ReLU with maximum \nth{4} degree are still of \nth{3} degree. Numbers in bold signify the best performing range for a given polynomial degree. Thick lines separate between polynomial degrees.}
		\midsepremove
		\begin{tabular}{clccccc}
			\cmidrule{1-6}\
			~ & ~ & \multicolumn{2}{|c|}{LeNet-5-like} & \multicolumn{2}{c}{SqueezeNet} \\ \cmidrule{2-6}
			~ & Nonlinearity & ReLU & ELU & ReLU & ELU \\ \cmidrule{1-6}\morecmidrules\cmidrule{1-6}
			Vanilla & Baseline 1 & 99.28\% & 99.01\%
			 & 76.35\% & 73.50\% & 1 \\ \cmidrule{1-6}\morecmidrules\cmidrule{1-6}
			 \multirow{6}{*}{\shortstack[c]{Min-Max\\([-1, 1])}}  & Baseline 2 & 99.55\% & 97.78\% & 81.87\% & 72.51\% & 2 \\ \cmidrule{2-6}
			~ & n=2 r=[-1, 1] & 96.65\% & 95.1\%    & 72.65\%   & 64.66\%  \\ \cmidrule{2-6}
			~ & n=3 r=[-1, 1] & -       & 97.45\%   & -         & 70.73\% \\ \cmidrule{2-6}
			~ & n=4 r=[-1, 1] & 99.33\% & 97.53\%   & 80.35\%   & 71.24\% \\ \cmidrule{2-6}
			~ & n=5 r=[-1, 1] & -       & 97.74\%   & -         & 72.19\% \\ \cmidrule{2-6}
			~ & n=6 r=[-1, 1] & 99.47\% & 97.76     & 81.23\%   & 72.23 \\ \cmidrule{1-6}
			\morecmidrules\cmidrule{1-6}
			 \multirow{11}{*}{\rotatebox[origin=c]{90}{Min-Max (range=[-2, 2])}}  & Baseline 2 & 98.75\% & 99.17\% & 80.79\% & 80.53\% & 3  \\ \cmidrule[1.5pt]{2-6}
			~ & n=2 r=[-1, 1] & \textbf{96.69\%}    & \textbf{98.51\%}  & 58.68\%           & \textbf{74.33\%} \\ \cmidrule{2-6}
			~ & n=2 r=[-2, 2] & 91.03\%             & 98.47\%           & \textbf{68.27\%}  & 71.42\% \\ \cmidrule[1.5pt]{2-6}
			~ & n=3 r=[-1, 1] & -                   & \textbf{99.03\%}  & -                 & 77.59\% \\ \cmidrule{2-6}
			~ & n=3 r=[-2, 2] & -                   & 98.88\%           & -                 & \textbf{77.86\%} \\ \cmidrule[1.5pt]{2-6} 
			~ & n=4 r=[-1, 1] & 7.96\%              & \textbf{99.18\%}  & 11.17\%           & 77.2\% \\ \cmidrule{2-6}
			~ & n=4 r=[-2, 2] & \textbf{98.41\%}    & 98.97\%           & \textbf{78.66\%}  & \textbf{79.43\%} \\ \cmidrule[1.5pt]{2-6} 
			~ & n=5 r=[-1, 1] & -                   & 93.49\%           & -                 & 18.54\% \\ \cmidrule{2-6}
			~ & n=5 r=[-2, 2] & -                   & \textbf{99.13\%}  & -                 & \textbf{80.11\%} \\ \cmidrule[1.5pt]{2-6} 
			~ & n=6 r=[-1, 1] & 19.4\%              & 94.84\%           & 9.87\%            & 19.14\% \\ \cmidrule{2-6}
			~ & n=6 r=[-2, 2] & \textbf{98.71\%}    & \textbf{99.11\%}  & \textbf{79.97\%}  & \textbf{80.17\%} \\ \cmidrule{1-6}
			\morecmidrules\cmidrule{1-6}
			\multirow{16}{*}{\rotatebox[origin=c]{90}{Min-Max (range=[-3,3])}}  & Baseline 2 & 99.49\% & 99.39\% & 82.04\% & 78.10\% & 4  \\ \cmidrule[1.5pt]{2-6}
			~ & n=2 r=[-1, 1] & 96.12\%         & \textbf{98.96\%}  & 33.12\%           & \textbf{73.30\%} \\ \cmidrule{2-6}
			~ & n=2 r=[-2, 2] & \textbf{99.18\%}& 98.91\%           & \textbf{75.75\%}  & 71.37\% \\ \cmidrule{2-6}
			~ & n=2 r=[-3, 3] & 97.11\%         & 98.68\%           & 71.36\%           & 68.42\% \\ \cmidrule[1.5pt]{2-6}
			~ & n=3 r=[-1, 1] & -               & 98.36\%           & -                 & 28.17\% \\ \cmidrule{2-6}
			~ & n=3 r=[-2, 2] & -               & \textbf{99.29\%}  & -                 & \textbf{77.76\%} & 5 \\ \cmidrule{2-6}
			~ & n=3 r=[-3, 3] & -               & 98.95\%           & -                 & 74.98\% \\ \cmidrule[1.5pt]{2-6}
			~ & n=4 r=[-1, 1] & 6.10\%          & 95.02\%           & 10.08\%           & 21.22\% \\ \cmidrule{2-6}
			~ & n=4 r=[-2, 2] & 99.18\%         & \textbf{99.34\%}  & 64.65\%           & \textbf{77.70\%} \\ \cmidrule{2-6}
			~ & n=4 r=[-3, 3] & \textbf{99.33\%}& 99.22\%           & \textbf{80.29\%}  & 77.11\% & 6  \\ \cmidrule[1.5pt]{2-6}
			~ & n=5 r=[-1, 1] & -               & 9.33\%            & -                 & 9.99\% \\ \cmidrule{2-6}
			~ & n=5 r=[-2, 2] & -               & 99.14\%           & -                 & 67.82\% \\ \cmidrule{2-6}
			~ & n=5 r=[-3, 3] & -               & \textbf{99.30\%}  & -                 & \textbf{77.79\%} \\ \cmidrule[1.5pt]{2-6}
			~ & n=6 r=[-1, 1] & 16\%            & 20.57\%           & 10.09\%           & 10.56\% \\ \cmidrule{2-6}
			~ & n=6 r=[-2, 2] & 93\%            & 99.22\%           & 18.94\%           & 68.36\% \\ \cmidrule{2-6}
			~ & n=6 r=[-3, 3] & \textbf{99.44\%}& \textbf{99.32\%}  & \textbf{81.46\%}  & \textbf{77.88\%} \\ \cmidrule{1-6}\
		\end{tabular}
		\midsepdefault
	\label{table:results}
	\end{table}
	
	\section{Discussion}\label{sec:discussion}
	The results of the experiments show that the proposed method can be used reliably for inference on homomorphically encrypted data. Here, we provide further interpretation of these results and give recommendations regarding the usage of our proposed method.
	
	\subsection{Min-Max normalization}\label{disc:minmax} 
\begin{figure}
		\centering
		\includegraphics[width=1\linewidth]{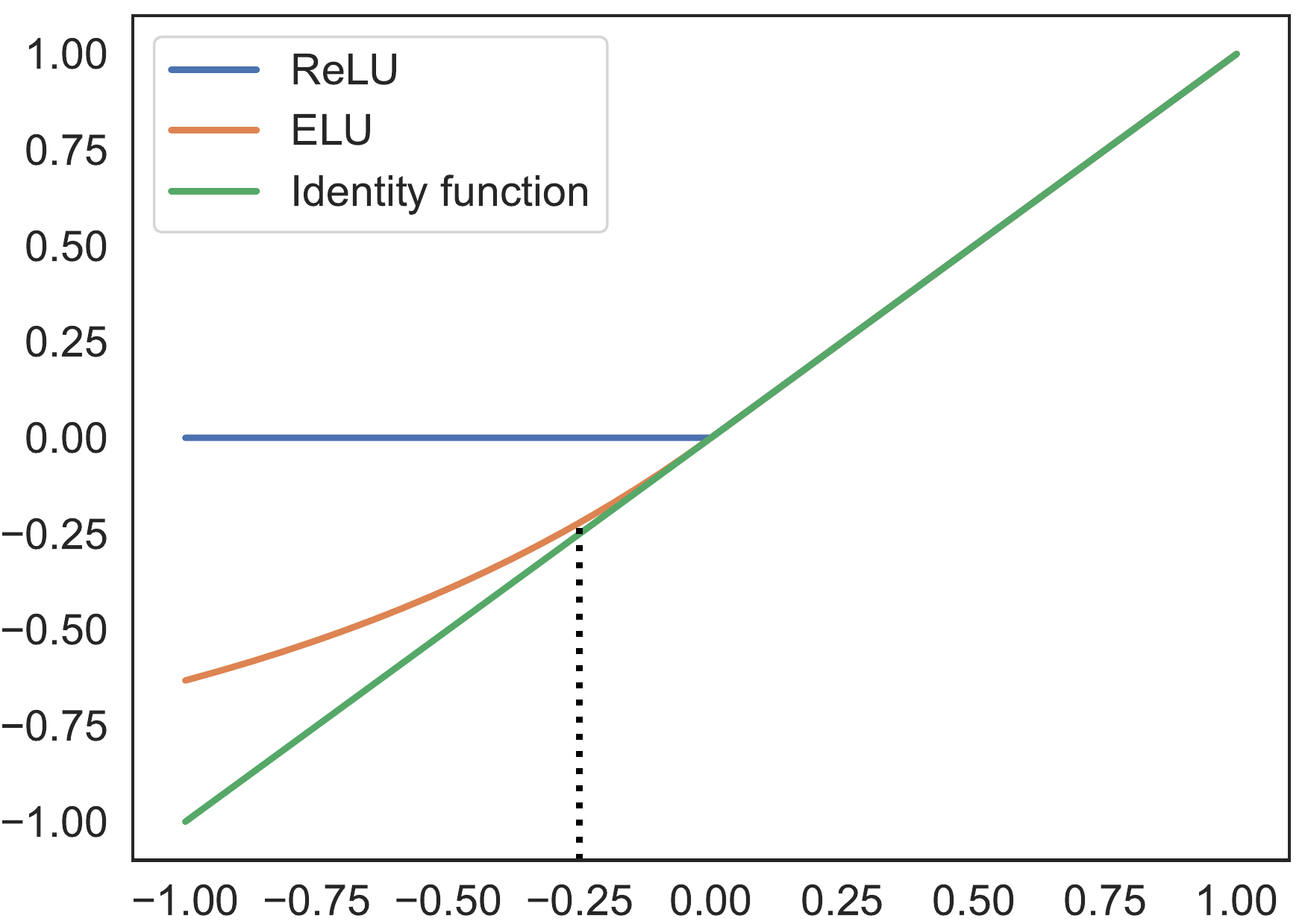}
		\caption{ELU and ReLU compared to the identity function in the range $[-1, 1]$. The vertical line at $-0.25$ marks the linear part of the ELU function. Notice that within the range $[-1, 1]$, most of the ELU function appears as linear.}
		\label{fig:linear_elu_relu}
	\end{figure}

	The Min-Max normalization layer offered an intuitive method to control the range of activations in the network by setting a single hyperparameter which explicitly defines the desired range. Its performance was consistently high, with different nonlinear functions and different approximations, using different polynomial degrees and approximation ranges. We attribute this to the Min-Max layer's ability to utilize the whole range of low polynomial approximation error, by scaling activations with a high absolute value to the target range. This is desirable, if we consider that the approximation error in different activations can have different signs (see absolute errors in Fig. \ref{fig:act_fn_abs_err}). Boura \etal \cite{BouraGG18} showed that using average instead of sum pooling layers yields more stability to perturbations in the network. Hence, average pooling in combination with opposite-signed error values leads to a higher chance of approximation errors to be canceled out. The shape of the approximated nonlinearity and the parameterization of the Min-Max layer plays an important role as well, as shown by significant performance drops if ELU is scaled to the range of $[-1, 1]$. In \autoref{fig:linear_elu_relu}, we show that constraining inputs to ELU to the range $[-1, 1]$ yields an almost linear function for most of the target range (approximately $[-0.25, 1]$), with the actual non-linearity lying only between $[-1, -0.25]$. This behavior greatly hinders the capacity of the network by turning it into a mostly linear function.
	
	\subsection{Nonlinear approximations} \label{disc: nonlinear_approx}
	As noted in \autoref{sec:experiments}, polynomial approximation functions with a matching polynomial degree and approximation range performed well in most cases, independent of the networks' depths. However, our results also show that the best performance is not necessarily achieved by approximation ranges that exactly match the target scaling range.
	
	 We explain this by the fact that for a given polynomial degree, the polynomial approximation of the nonlinearity gets more accurate if a smaller approximation range is considered. Since the activation values have a Gaussian-like distribution roughly centered at 0 after the Min-Max layer (\autoref{fig:activations_dists}), most of the activations benefit from a very low approximation error. However, as the polynomial degree increases, activations that lie outside of the permitted range lead to a polynomially high approximation error. The consequence is a noticeable drop of network performance. This is visualized in \autoref{fig:error_dist}: among the three approximation ranges shown, the approximation function over the range $[-1, 1]$ has the lowest error in this interval, which covers approximately $63\%$ of input activation samples. In comparison, the approximation over the range $[-2, 2]$ has a slightly higher approximation error in the interval $[-1, 1]$. However, up until the range $[-2, 2]$, the errors are smaller, yielding relatively low approximation errors for approximately $97\%$ of input activation samples.
	
	Empirical results show that ELU can preserve the prediction accuracies with a lower degree polynomial approximations. We attribute this property to two factors: 
	\begin{enumerate}
	    \item The region with highest density of activations lies around 0. It can be approximated in ELU with a lower degree than in ReLU, because of the lack of a discontinuity at the origin. Hence, less approximation capacity is needed, and part of that capacity can be used to approximate a bigger range (cf. \autoref{fig:act_fn_abs_err} for the approximation errors of polynomial of the same degree).
	    \item The polynomial approximation of ELU does not deviate too much away from the ground truth outside the approximated range in contrast to ReLU. This is visualized in \autoref{fig:fn_approximations}.
	\end{enumerate}
	As expected, the performances of both ReLU and ELU approximations get closer to the baseline~2, as the approximation degree increases. We stop at degree 6, as higher degrees will not be practical in the context of LHE due to the magnitude of the noise that will be generated because of the depth of the multiplicative circuit.
    \subsection{Design recommendations for architects}
    To utilize the full potential of the Min-Max layer, we highlight several factors that a network architect has to consider at design time, and we offer several guidelines on how to set hyperparameters of the Min-Max layers and of the approximations of the nonlinear functions accordingly.
    \begin{itemize}
        \item \textit{The original nonlinear function:} the function's shape puts a hard constraint on the target scaling range of the Min-Max layer. The designer needs to take care that the range is big enough to preserve the nonlinearity property of that function. For example, as discussed, approximating ELU with the target scaling range of $[-1, 1]$ is suboptimal since the resulting function is linear in a majority of the input range (cf. \autoref{disc:minmax}).
        \item \textit{Depth of the network:} As the network gets deeper, the architect has to choose a polynomial of lower degree to minimize the noise generation in the LHE ciphertexts. Additionally, to maintain the highest possible fidelity to the baseline~2 performance, the approximation range has to decrease, to match the approximation capacity of that degree (cf. discussion in \autoref{disc: nonlinear_approx}).
        \item \textit{Hybrid polynomial degrees:} Cheney \etal \cite{CheneySK17} showed that the last layers of a network are relatively resilient to approximation errors, compared to earlier layers. The architect can employ a polynomial approximation of high degree at the early layers, and a lower degree in the subsequent layers in order to increase the performance with minimal increase in noise generation in the ciphertexts. We demonstrate this with SqueezeNet with a target scaling range of $[-2, 2]$ by employing a polynomial of \nth{5} degree in the first three blocks then a polynomial of \nth{2} degree in the remaining six blocks, The prediction accuracy increased 6.2 percentage points compared to a network with only polynomials of degree 2.  
    \end{itemize}
    
	\section{Conclusion}
	Min-Max normalization has been one of the preprocessing methods used in the literature and the industry. In this work, a novel use within the hidden layers was presented to facilitate DNN inference on homomorphically encrypted data. The proposed method is characterized by 
    \begin{enumerate}
        \item \textit{Ease of initialization and trivial setting of hyperparameters:} the proposed method only has a single hyperparameter that explicitly defines the desired input range to the nonlinear functions.
        \item \textit{Simplicity of the training process:} No additional training steps are required other than the traditional training process with plain data.
        \item \textit{Compatibility:} Min-max normalization requires little change to current common architectures.
    \end{enumerate}
	In the context of DNN with LHE, this work proposes a solution to the nonlinearity problem by allowing the network architect to use try-and-tested nonlinear functions during training, and to use polynomial function approximations at inference time, with minimum loss in performance.

    \section*{Acknowledgments}
    The study was supported in parts by the German Federal Ministry of Education and Research (BMBF) in connection with the foundation of the German Center for Vertigo and Balance Disorders (DSGZ) (grant number 01 EO 0901).

\ifCLASSOPTIONcaptionsoff
  \newpage
\fi


\bibliographystyle{IEEEtran}
\bibliography{minmax}

%








\end{document}